\documentclass[letterpaper, 10 pt, conference]{ieeeconf}
\usepackage{times}
\usepackage{graphicx}
\usepackage{subcaption}
\usepackage{todonotes}
\usepackage{float}
\usepackage{tikz}
\usepackage{pgfplots}
\usepackage{tabularx}
\usepackage{hyperref}
\hypersetup{
    colorlinks=true,
    linkcolor=blue,
    filecolor=magenta,
    urlcolor=cyan,
    }

\usepackage{adjustbox}
\usepackage{subcaption}

\pgfplotsset{compat=1.17}

\title{\LARGE \bf Robot Operating System 2: Design, Architecture, and Uses In The Wild}

\makeatletter
\newcommand{\linebreakand}{%
  \end{@IEEEauthorhalign}
  \hfill\mbox{}\par
  \mbox{}\hfill\begin{@IEEEauthorhalign}
}
\makeatother

\author{
\authorblockN{Steve Macenski}
\authorblockA{\textit{Samsung Research America}\\
s.macenski@samsung.com}
\and
\authorblockN{Tully Foote}
\authorblockA{\textit{Open Robotics}\\
tfoote@openrobotics.org}
\and 
\authorblockN{Brian Gerkey}
\authorblockA{\textit{Open Robotics}\\
gerkey@openrobotics.org}
\and 
\linebreakand % See macro above for doing this
\authorblockN{Chris Lalancette}
\authorblockA{\textit{Open Robotics}\\
clalancette@openrobotics.org}
\and 
\authorblockN{William Woodall}
\authorblockA{\textit{Open Robotics}\\
william@openrobotics.org}
}

\begin{document} 
\maketitle

\begin{abstract}

\footnote{This manuscript has been accepted for publication in Science Robotics. This version has not undergone final editing. Please refer to the complete version of record at \url{https://www.science.org/doi/10.1126/scirobotics.abm6074}. The manuscript may not be reproduced or used in any manner that does not fall within the fair use provisions of the Copyright Act without the prior, written permission of AAAS.}The next chapter of the robotics revolution is well underway with the deployment of robots for a broad range of commercial use-cases.
Even in a myriad of applications and environments, there exists a common vocabulary of components that robots share - the need for a modular, scalable, and reliable architecture; sensing; planning; mobility; and autonomy.
The Robot Operating System (ROS) was an integral part of the last chapter, demonstrably expediting robotics research with freely-available components and a modular framework.
However, ROS 1 was not designed with many necessary production-grade features and algorithms.
ROS~2 and its related projects have been redesigned from the ground up to meet the challenges set forth by modern robotic systems in new and exploratory domains at all scales.
In this review, we highlight the philosophical and architectural changes of ROS~2  powering this new chapter in the robotics revolution.
We also show through case studies the influence ROS~2 and its adoption has had on accelerating real robot systems to reliable deployment in an assortment of challenging environments.

\end{abstract}

\section{Introduction}
% general middleware/robotics framework introduction.
Many software platforms have been proposed, sometimes called {\em middlewares}, introducing modular and adaptable features making it easier to build robot systems.
Over time, some middlewares have grown to become rich ecosystems of utilities, algorithms, and sample applications.
Few rival the Robot Operating System (ROS 1) in its significance on the maturing robotics industry.

% intro ROS, history and transition into ROS2
ROS~1 was popularized by the robotics incubator, Willow Garage \cite{ros2009icra}.
Every effort was made to create a quality and performant system, but security, network topology, and system up-time were not prioritized.
Regardless, ROS~1 has become influential in nearly every intelligent machine sector.
Its commercial rise was the result of flagship projects providing autonomous navigation, simulation, visualization, control, and more \cite{Chitta2017, 5509725,moveit}.
As commercial opportunities transitioned into products, ROS's foundation as a research platform began to show its limitations. 
Security, reliability in non-traditional environments, and support for large scale embedded systems became essential to push the industry forward.
Further, many companies where building workarounds on top or inside of ROS~1 to create reliable applications \cite{Cairl2018}.

% ROS2
The second generation of the Robot Operating System, ROS~2, was redesigned from the ground up to address these challenges while building on the success of its community-driven capabilities \cite{macenski2020marathon2}.
ROS~2 is based on the Data Distribution Service (DDS), an open standard for communications that is used in critical infrastructure such as military, spacecraft, and financial systems \cite{1203555}.
It solves many of the problems in building reliable robotics systems.
DDS enables ROS~2 to obtain best-in-class security, embedded and real-time support, multi-robot communication, and operations in non-ideal networking environments.
DDS was selected after considering other communication technologies, e.g. ZeroMQ, RabbitMQ, due to its breadth of features including a UDP transport, distributed discovery, a builtin security standard \cite{ros2-on-dds}.

% With the background in mind, what's this paper about?
In this review article, we will establish ROS~2's state of the art suitability for modern robot systems and showcase the technological and philosophical changes that have driven its success.
Then, we will expand on that foundation to demonstrate how ROS~2 is influencing the deployment of autonomous systems in several unique domains.
Five case studies explore how ROS~2 has enabled or accelerated robots into the wild on land, sea, air, and even space.

\section{Related Work}
% demonstrate we know what came before and lay foundation for the novelty of the thing we're presenting
% For each: describe what they do, when they were made, requirements, the goods, the bads, what is in their ecosystems (is there an ecosystem)?
% reference papers for these things, think about ~1/2 page introduction for each

The history of robot software is long and storied, going back more than 50 years with robots like Shakey~\cite{kuipers2017}.
Over time, much has been written about how to structure classical planners, concurrent behaviors, and three-layer architectures~\cite{fikes1971strips,brooks1986robust,gat1998three}.
An early example of this is the Task Control Architecture (TCA), which was used to control a variety of robots.
For example, CARMEN was built on TCA's message-passing system called IPC (Inter-Process Communications) \cite{simmons1994structured,montemerlo2003perspectives}.
Message-passing has its own rich history in distributed systems: from IBM's work on {\em message queuing}, Java's Jini, and middlewares such as MQTT \cite{mohan1994recent,waldo1999jini,mqtt2019}.

Robotics frameworks provide architectural methods to decompose complex software into smaller and more manageable pieces.
Some of these components can find re-use in other systems and may be established into libraries to be leveraged by users.
An early attempt to manage this complexity was via a client/server approach in Player \cite{player2001iros}.
A Player server communicates with robot hardware and runs the algorithms needed to perform its task.
Clients can connect to the server to extract data and control the robot over a TCP connection.
However, its architecture hampered reliability, code reuse, and ability to change out components.

YARP aids in building control systems organized as peers, communicating over several protocols \cite{yarp2006ars}.
It facilitates research development and collaboration by promoting code reuse and modularity while retaining high performance.
YARP can be used for any application, but its community has focused on humanoid and legged robotics, such as iCub and the MIT Cheetah and only supports C++.

LCM is a middleware which uses a publish/subscribe model with bindings in many languages.
It concentrates on handling messaging and data marshalling in high-bandwidth low-latency environments \cite{lcm2010iros}.
This limits the range of robotic applications for which LCM can be effectively used.
OROCOS is a set of libraries for robot control, focused on real-time control systems and related topics, such as computing kinematic chains and Bayesian filtering \cite{orocos2003icra}.
The project has grown into a full framework integrating the CORBA middleware and tooling for deterministic computation in real-time applications.
The LCM and OROCOS frameworks each concentrate on smaller pieces of the overall system, with a non-trivial proportion of the overall robotics problem left to the end-user.

ROS 1 contains a set of libraries that are useful when building many kinds of robots \cite{ros2009icra}.
There are utilities for monitoring processes, introspecting communications, receiving time-series transformations, and more.
ROS 1 also has a large ecosystem of sensor, control, and algorithmic packages made available by community contributions, enabling a small team to build complex robotics applications.
While ROS 1 solves many of the complexity issues inherent to robotics, it struggles to consistently deliver data over lossy links (like WiFi or satellite links), has a single point of failure, and does not have any built-in security mechanisms.
A table of key differences between ROS~1 and ROS~2 can be seen in Table \ref{tab:ro2features}.

The ROS~1 community attempted to address some of these concerns, but in nearly all cases, there were compromises made due architectural and engineering limitations.
For example, to address the single point of failure (``rosmaster''), it was required to patch all of the existing client libraries individually with bespoke solutions.
In other cases, it was possible to extend ROS~1 for security, via the SROS project.
Though successful, it was difficult to maintain and needed further development to meet security trends.
These are just two of the many attempts to patch ROS~1, which extended its useful lifetime but did not solve its core limitations.

\begin{table}[]
    \centering
    \begin{tabularx}{\columnwidth}{|l|>{\raggedright\arraybackslash}X|>{\raggedright\arraybackslash}X|}
    \hline
    {\bf Category} & {\bf ROS 1} & {\bf ROS 2} \\
    \hline
{\bf Network Transport} & Bespoke protocol built on TCP/UDP & Existing standard (DDS), with abstraction supporting addition of others \\
\hline
{\bf Network Architecture} & Central name server ({\tt roscore}) & Peer-to-peer discovery \\
\hline
{\bf Platform Support} & Linux & Linux, Windows, macOS\\
\hline
{\bf Client Libraries} & Written independently in each language & Sharing a common underlying C library ({\tt rcl})\\
\hline
{\bf Node vs. Process} & Single node per process & Multiple nodes per process\\
\hline
{\bf Threading Model} & Callback queues and handlers & Swappable executor\\
\hline
{\bf Node State Management} & None & Lifecycle nodes\\
\hline
{\bf Embedded Systems} & Minimal experimental support ({\tt rosserial}) & Commercially supported implementation (micro-ROS)\\
\hline
{\bf Parameter Access} & Auxilliary protocol built on XMLRPC & Implemented using service calls\\
\hline
{\bf Parameter Types} & Type inferred when assigned & Type declared and enforced\\
\hline
    
    \end{tabularx}
    \caption{Summary of ROS 2 features compared to ROS 1}
    \label{tab:ro2features}
\end{table}

\section{ROS~2}

ROS~2 is a software platform for developing robotics applications, also known as a robotics software development kit (SDK).
Importantly, ROS~2 is {\em open source}.
ROS~2 is distributed under the Apache 2.0 License, which grants users broad rights to modify, apply, and redistribute the software, with no obligation to contribute back \cite{apache2license}.
ROS~2 relies on a {\em federated} ecosystem, in which contributors are encouraged to create and release their own software.
Most additional packages also use the Apache 2.0 License or similar.
Making code {\em free} is fundamental to driving mass adoption - it allows users to leverage ROS~2 without constraining how they use or distribute their applications.

\subsection{Scope}
ROS~2 supports a broad range of robotics applications, from education and research to product development and deployment.
It comprises a large set of interrelated software components that are commonly used to develop robotics applications. The software ecosystem is divided into three categories:
\begin{itemize}
    \item {\bf Middleware:} Referred to as the {\em plumbing}, the ROS~2 middleware encompasses communication among components, from network APIs to message parsers.
    \item {\bf Algorithms:} ROS~2 provides many of the algorithms commonly used when building robotics applications, e.g. perception, SLAM, planning, and beyond.
    \item {\bf Developer tools:} ROS~2 includes a suite of command-line and graphical tools for configuration, launch, introspection, visualization, debugging, simulation, and logging.
    There is also a large suite of tools for source management, build processes, and distribution.
\end{itemize}
In this section, we will explore the first category, the middleware, as the foundation of ROS~2.

\subsection{Design} \label{design}

\subsubsection{Design Principles} \label{sec:principles}

The design of ROS~2 has been guided by a set of principles and a set of specific requirements.
The following principles are asserted:

\textbf{Distribution} As with similarly complex domains, problems in robotics are best tackled with a distributed systems approach \cite{Birman1987ExploitingVS}.
Requirements are separated into functionally independent components, like device drivers for hardware, perception systems, control systems, executives, and so on.
At run-time, these components have their own execution context and share data via explicit communication.
This composition should be conducted in a decentralized and secure manner.

\textbf{Abstraction} To govern communication, interface specifications must be establish.
These messages define the semantics of the data exchanged.
A favorable abstraction balances the benefits of exposing the details of a component against the costs of overfitting the rest of the application to that component, thereby making it difficult to substitute an alternative.
This approach leads to an ecosystem of interoperable components abstracted away from specific vendors of hardware or software components \cite{corbet2005linux}.

\textbf{Asynchrony} The messages defined are communicated among the components asynchronously, creating an event-based system \cite{muhl2006distributed}.
With this approach, an application can work across the multiple time domains that arise from combining physical devices with a host of software components; each of which may have its own frequency for providing data, accepting commands, or signaling events.

\textbf{Modularity} The UNIX design goal to `make each program do one thing well' is mirrored \cite{McIlroy78}.
Modularity is enforced at multiple levels, across library APIs, message definitions, command-line tools, and even the software ecosystem itself.
The ecosystem is organized into a large number of federated packages, as opposed to a single codebase.

\vspace{1em} \noindent
We do not pretend that these design principles are universal and without trade-offs.
Asynchrony can also make it more difficult to achieve deterministic execution.
For any single, well-defined problem, it is possible to construct a special-purpose monolithic solution that is more computationally efficient because it does not involve abstractions or distributed communication.

However, after a decade of experience with the ROS~1 project, we claim that adherence to these principles will generally lead to better outcomes.
This approach facilitates code reuse, software testing, fault isolation, collaboration within interdisciplinary project teams, and cooperation at a global scale.

\subsubsection{Design Requirements}
ROS~2 aims to meet certain requirements based on the design principles and needs of robotics developers.

\textbf{Security} Any software that interacts with a network must include features to secure that interaction against accidental and malicious misuse. ROS~2's integrated security system includes authentication, encryption, and access control \cite{Gerard2017,ros2-dds-security,ddssecurity}. Designers can configure ROS~2 to meet their needs through access control policies that define who can communicate about what \cite{White2018}.

\textbf{Embedded systems} As a general rule, a robot includes sensors, actuators, and other peripherals. These devices can be relatively sophisticated, containing micro-controllers that need to communicate with CPU(s) where ROS~2 is running. A full ROS~2 stack is not expected to run on small embedded devices, though ROS~2 should facilitate and standardize integration of CPUs and micro-controllers. {\em Micro-ROS} allows ROS~2 to be reused on embedded systems \cite{Lutkebohle2019}.

\textbf{Diverse networks} Robots are used in a variety of networking environments, from wired LAN for robot arms on assembly lines to multi-hop satellite connections for planetary rovers. Additionally, robots will often use internal networks to connect processes within and across CPUs. ROS~2 provides {\em quality of service} that configures how data flows through the system, thereby adapting to the constraints of a network \cite{dds-spec}.

\textbf{Real-time computing} From humanoids to self-driving cars, it is common for robot applications to include real-time computing requirements. To meet safety and/or performance goals, some parts of a system must execute in deterministic amounts of time. ROS~2 offers APIs for developers of real-time systems to enforce application-specific constraints \cite{puck2020distributed, staschulat2020rclc}.

\textbf{Product readiness} When a robot moves beyond the lab and into commercial use, new constraints are introduced. ROS~2 aims to meet product requirements spanning design, development, and project governance. One objective result of these efforts is Apex.AI's functional safety (ISO 26262) certification of their ROS~2-based autonomous vehicle software \cite{apex-cert}. This allows ROS~2 to be run in safety critical systems like autonomous vehicles and heavy machinery.

\subsection{Communication Patterns}
The ROS~2 APIs provide access to communication patterns.
These are notably topics, services, and actions, which are organized under the concept of a {\em node}.
ROS~2 also provides APIs for parameters, timers, launch, and other auxiliary tools which can be used to design a robotic system.

\begin{figure*}[ht]
    \centering
    \includegraphics[width=0.7\textwidth,keepaspectratio]{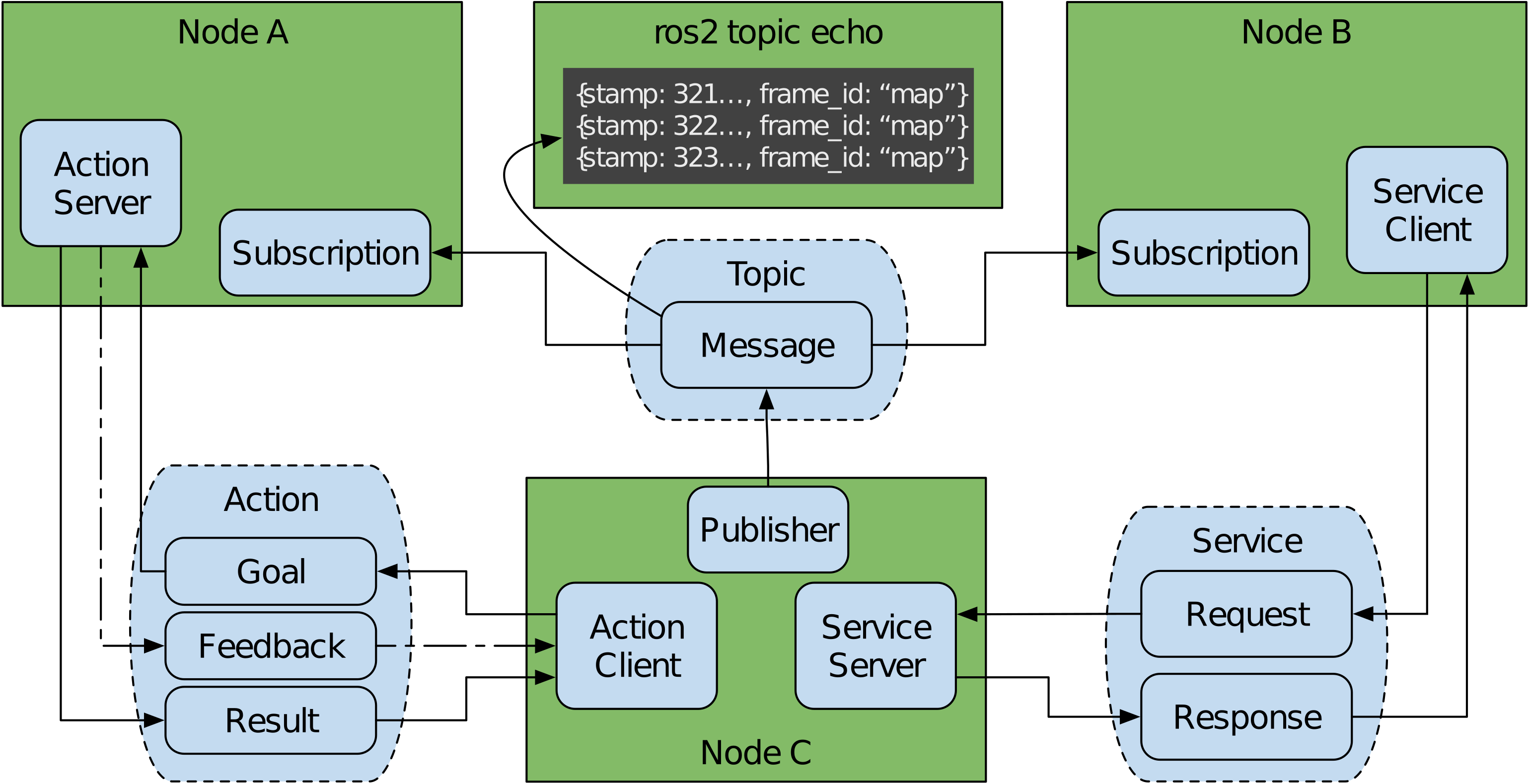}
    \caption{ROS~2 node interfaces: topics, services, and actions.}
    \label{fig:topics}
\end{figure*}

\textbf{Topics}
The most common pattern that users will interact with is {\em topics}, which are an asynchronous message passing framework.
This is similar to other asynchronous frameworks, such as ASIO \cite{asio}.
ROS~2 provides the same publish-subscribe functionality, but focuses on using asynchronous messaging to organize a system using strongly typed interfaces.
It does so by organizing endpoints in a computational graph under the concept of a {\em node}.
The node is an important organizational unit which allows a user to reason about a complex system, shown in Fig. \ref{fig:topics}.

The anonymous publish-subscribe architecture allows many-to-many communication, which is advantageous for system introspection.
A developer may observe any messages passing on a topic by creating a subscription to that topic without any changes.

\textbf{Services}
Asynchronous communication is not always the right tool.
ROS~2 also provides a request-response style pattern, known as {\em services}.
Request-response communication provides easy data association between a request and response pair, which can be useful when ensuring a task was completed or received, shown in Fig. \ref{fig:topics}.
Uniquely, ROS~2 allows a service client's process to not be blocked during a call.
Services are also organized under a node for organization and introspection, allowing a subsystem's interfaces to appear together in system diagnostics.

\textbf{Actions}
A unique communication pattern of ROS~2 is the {\em action}.
Actions are goal-oriented and asynchronous communication interfaces with a request, response, periodic feedback, and the ability to be canceled, Fig. \ref{fig:topics}.
This pattern is used in long-running tasks like autonomous navigation or manipulation, though it has a variety of uses.
Like services, actions are non-blocking and organized under the node.

\subsection{Middleware Architecture}

Adhering to the previous design philosophies, the architecture of ROS~2 consists of several important abstraction layers distributed across many decoupled packages.
These abstraction layers make it possible to have multiple solutions for required functionality, e.g. multiple middlewares or loggers.
Additionally, the distribution across many packages allows users to replace components or take only the pieces of the system they require, which may be important for certification.

\begin{figure*}[ht]
    \centering
    \includegraphics[width=0.7\textwidth,keepaspectratio]{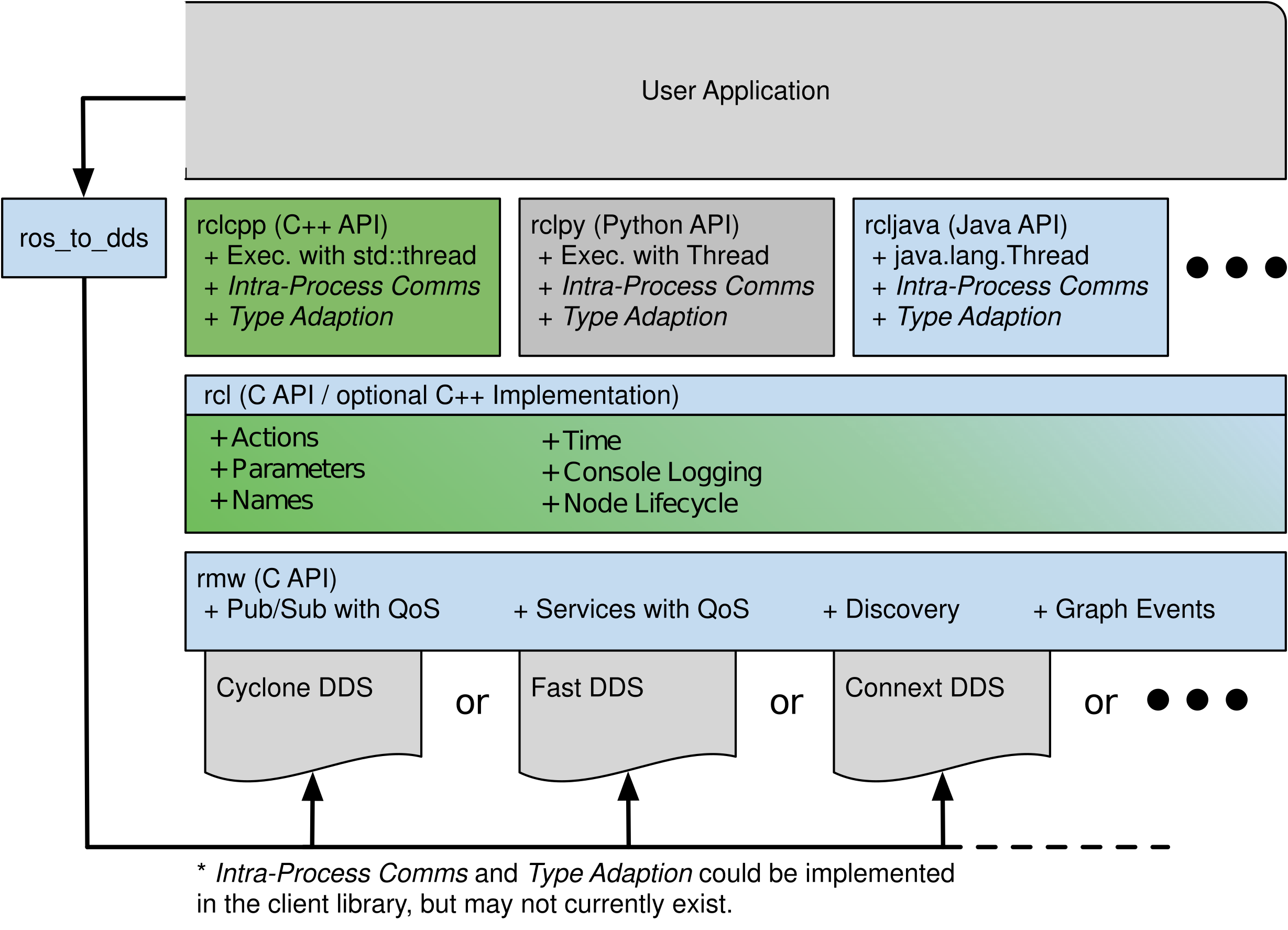}
    \caption{ROS~2 Client Library API Stack.}
    \label{fig:rosclientlibraryapistack}
\end{figure*}

\subsubsection{Abstraction Layers} \label{abstractions}

Figure \ref{fig:rosclientlibraryapistack} displays the abstraction layers within ROS~2.
They are generally hidden behind the client library during development and developers would only need to be aware of them for unusually application-specific needs.
Most users will experience {\em only} the client libraries.

The client libraries provide access to the core communication APIs.
They are tailored to each programming language to make them more idiomatic and take advantage of language specific features.
Communication is agnostic to how the system is distributed across compute resources - whether they be in the same process, a different process, or even a different computer.
A user may distribute their application across multiple machines and processes, and even leverage cloud compute resources, with minimal changes to the source code.
ROS~2 is capable of connecting to cloud resources over the internet.
There are products to assist integration of ROS~2 into cloud platforms, such as AWS IoT RoboRunner and related RoboMaker products \cite{aws-iot-roborunner}.
It is advisable, however, to use more specialized purpose-built technologies.

The client libraries depend on an intermediate interface, {\tt rcl}, which provides common functionality to each client library.
This library is written in C and is used by all of the client libraries, although not required.
Below {\tt rcl}, the middleware abstraction layer called {\tt rmw} (ROS MiddleWare) provides the essential communication interfaces.
The vendors for each middleware implements the {\tt rmw} interface and are made interchangeable without code changes.

Users may choose different {\em rmw} implementations, and thereby different middleware technologies, based on a variety of constraints like performance, software license, or supported platforms.
While all of the supported {\em rmw}s are based on DDS, a few community-supported {\em rmw}s exist for other communications methods \cite{1203555}.
This abstraction layer provides flexibility to ROS~2, allowing it to change over time with minimal impact to the systems built atop it.

The network interfaces (e.g. topics, services, actions) are defined with {\em Message Types} using an {\em Interface Description Language} (IDL).
ROS~2 defines these types using the ros idl format ({\tt .msg} files) or the OMG IDL standard ({\tt .idl} files).
User-provided interface definitions are generated at compile time and create code required for communication in any client library language.

\subsubsection{Architectural Node Patterns}

There are additional architectural patterns to help developers structure their programs.
ROS~2 provides a pattern for managing the lifecycle of {\em nodes} which transition through a state machine with states like Unconfigured, Inactive, Active, and Finalized.
These states allow system integrators to control when certain {\em nodes} are active.
This is an important tool for coordinating various parts of the distributed asynchronous system.

As previously discussed in the last section, communications are agnostic to the location of endpoints within machines and processes.
However, in which machine or process to put each node is not something that should be decided when writing the node, but instead depends on how the node is used in the larger system.
Nodes that are written as {\em components} can be allocated to any process as a configuration.
This is an important feature for systems under development, allowing the developer to rearrange where {\em nodes} are running based on a variety of circumstances.
For example, multiple {\em nodes} might be configured to share a process in order to conserve system resources or reduce latency.

\subsection{Software Quality}
% OUTLINE DOC
%   - Quality (provable claims)
%       - design articles
%       - quality level process
%       - Certification compliance processes
%       - Linting / static analysis / CI
%       - High test coverage

For ROS~2 to be adopted in critical applications, it must be designed and implemented in a demonstrably high-quality manner. Regulatory and  certification bodies need to understand the current state of a system and the processes that led to it. To that end, a three-part approach is continuously executed to measure and expose software quality:
\begin{itemize}
    \item {\bf Design documentation:} Prior to a major addition, a written rationale and design for the work must be established. This documentation manifests as a design article or a ROS Enhancement Proposal (REP) \cite{ros2design-site, reps-site}. At the time of writing, there are 44 design articles and 7 REPs documenting the design of ROS~2.
    \item {\bf Testing:} Each feature in ROS~2 requires tests to ensure that it behaves correctly. Those tests are executed regularly in continuous integration. A combination of unit and integration tests are deployed, as well as a suite of static analysis tools (``linters''). At the time of writing, 32,000--33,000 tests are run on ROS~2, including 13 linters.
    \item {\bf Quality declaration:} Not every ROS~2 package needs to be rigorously documented and tested. Thus, a multi-level quality policy is defined \cite{rep-2004}. This policy defines the requirements for each quality level in terms of development practices, test coverage, security, and more. At the time of writing, 45 ROS~2 packages have achieved the highest level, Quality Level 1.
\end{itemize}

\subsection{Performance and Reliability}
% OUTLINE DOC
%   - [moved from Performance section] Reliability (provable claims)
%       - Based on DDS: ready for mission critical use-cases and multiple options for the best for your situation

Networking is an important aspect of robotics frameworks.
In reliable networking situations, the standard solution is TCP/IP due to its optimizations in most operating systems.
Unfortunately, TCP/IP struggles to deliver data in wireless communications since interruptions can cause back-offs, re-transmits, and delays.
ROS~1 was built on TCP/IP and suffered in these situations.

ROS~2 does not struggle in these situations.
DDS uses UDP to deliver data, which does not attempt to re-transmit data.
Instead, DDS decides when and how to re-transmit in unreliable conditions.
DDS introduces Quality-of-Service (QoS) to expose these settings to optimize for the available bandwidth and latency.

The {\em reliability} setting determines whether message delivery is guaranteed.
Using `best-effort', the publisher will attempt to deliver the message once, useful when new data will make the old obsolete (e.g. sensor data).
Set to `reliable', the publisher will continue to send data until the receiver acknowledges receipt.

The {\em durability} QoS setting determines the persistence of a message.
`Volatile' messages will be forgotten once after being sent.
Meanwhile, `transient-local' will store and send late-joining subscriptions data as necessary.

A connection's {\em history} determines the behavior when the network cannot keep up with the data.
Set to `keep-all', all data is retained until the application consumes it.
Most applications use `keep-last', which retains a fixed-sized queue of data, overriding the oldest as needed.
Other settings, including {\em deadline}, {\em lifespan}, {\em liveliness}, and {\em lease duration} to help in designing real-time systems.

\begin{figure*}[ht]
\centering
\subcaptionbox{Different message sizes\label{fig:size}}
{
\begin{subfigure}[b]{0.3\textwidth}
\begin{adjustbox}{width=\linewidth}
\begin{tikzpicture}[baseline]
\begin{semilogxaxis}[
    xlabel={Message Size (bytes)},
    ylabel={Mean Latency (ms)},
    xmax=8500,
    ymin=0, ymax=10,
    xtick={1,4,16,32,60,512,1024,2048,4096,8192},
    xticklabels={1k,4k,16k,32k,60k,512k,1m,2m,4m,8m},
    ytick={0,2,4,6,8,10},
    legend pos=north west,
    legend cell align={left},
    grid=both,
    grid style=dashed,
    tick label style={font=\tiny}
]

\addplot[
    color=blue,
    mark=square,
    ]
    coordinates {
    (1,0.08347199999999999)(4,0.08571133333333333)(16,0.08608933333333334)(32,0.08514933333333331)(60,0.094492)(512,0.15904666666666667)(1024,0.22696666666666676)(2048,0.36730999999999997)(4096,0.39932333333333336)(8192,1.2045666666666668)
    };
\addlegendentry{Intra-process}

\addplot[
    color=orange,
    mark=square,
    ]
    coordinates {
    (1,0.093669)(4,0.08464066666666664)(16,0.086858)(32,0.10848666666666666)(60,0.12359999999999996)(512,0.33722)(1024,0.3401266666666668)(2048,0.5068400000000001)(4096,0.9455566666666666)(8192,3.578533333333333)
    };
\addlegendentry{Single process}

\addplot[
    color=red,
    mark=square,
    ]
    coordinates {
    (1,0.19110000000000002)(4,0.17738666666666666)(16,0.19861666666666666)(32,0.20372333333333334)(60,0.2655833333333334)(512,0.44927000000000006)(1024,0.7091266666666667)(2048,1.3768000000000002)(4096,3.7576000000000005)(8192,8.366433333333331)
    };
\addlegendentry{Inter-process}
    
\end{semilogxaxis}
\end{tikzpicture}
\end{adjustbox}
\end{subfigure}
\begin{subfigure}[b]{0.3\textwidth}
\begin{adjustbox}{width=\linewidth}
\begin{tikzpicture}[baseline]
\begin{semilogxaxis}[
    xlabel={Message Size (bytes)},
    ylabel={Mean CPU (\%)},
    xmax=8500,
    ymin=0, ymax=35,
    xtick={1,4,16,32,60,512,1024,2048,4096,8192},
    xticklabels={1k,4k,16k,32k,60k,512k,1m,2m,4m,8m},
    ytick={0,5,10,15,20,25,30,35},
    legend pos=north west,
    legend cell align={left},
    grid=both,
    grid style=dashed,
    tick label style={font=\tiny}
]

\addplot[
    color=blue,
    mark=square,
    ]
    coordinates {
    (1,2.1751)(4,2.2361666666666675)(16,2.1473333333333335)(32,2.2306666666666675)(60,2.3803333333333336)(512,3.4512333333333323)(1024,4.516966666666667)(2048,6.653033333333331)(4096,7.103533333333331)(8192,17.642999999999997)
    };
\addlegendentry{Intra-process}

\addplot[
    color=orange,
    mark=square,
    ]
    coordinates {
    (1,2.3131333333333344)(4,2.1030333333333333)(16,2.1363)(32,2.5635333333333334)(60,2.8299333333333343)(512,6.3589666666666655)(1024,6.072166666666665)(2048,8.729166666666664)(4096,16.065)(8192,30.16499999999999)
    };

\addlegendentry{Single process}

\addplot[
    color=red,
    mark=square,
    ]
    coordinates {
    (1,4.10224)(4,3.82456)(16,4.26262)(32,4.47352)(60,5.6937)(512,10.29186)(1024,16.24946)(2048,30.7153)(4096,32.9299)(8192,34.1253)
    };

\addlegendentry{Inter-process}

\end{semilogxaxis}
\end{tikzpicture}
\end{adjustbox}
\end{subfigure}
\begin{subfigure}[b]{0.31\textwidth}
\begin{adjustbox}{width=\linewidth}
\begin{tikzpicture}[baseline]
\begin{semilogxaxis}[
    xlabel={Message Size (bytes)},
    ylabel={Sending rate (Hz)},
    xmax=8500,
    ymin=0, ymax=1000,
    xtick={1,4,16,32,60,512,1024,2048,4096,8192},
    xticklabels={1k,4k,16k,32k,60k,512k,1m,2m,4m,8m},
    ytick={0,100,200,300,400,500,600,700,800,900,1000},
    legend pos=south west,
    legend cell align={left},
    grid=both,
    grid style=dashed,
    tick label style={font=\tiny}
]

\addplot[
    color=blue,
    mark=square,
    ]
    coordinates {
    (1,999.5666666666667)(4,999.5333333333333)(16,999.6)(32,999.6333333333333)(60,999.5333333333333)(512,999.6333333333333)(1024,999.5)(2048,999.4666666666667)(4096,999.4333333333333)(8192,862.9333333333333)
    };
\addlegendentry{Intra-process}

\addplot[
    color=orange,
    mark=square,
    ]
    coordinates {
    (1,999.5666666666667)(4,999.6)(16,999.5666666666667)(32,999.5666666666667)(60,999.5)(512,999.5666666666667)(1024,999.5333333333333)(2048,999.5333333333333)(4096,999.5)(8192,553.2)
    };
\addlegendentry{Single process}

\addplot[
    color=red,
    mark=square,
    ]
    coordinates {
    (1,999.2333333333333)(4,999.2666666666667)(16,999.2333333333333)(32,999.2333333333333)(60,999.2666666666667)(512,999.2)(1024,999.2333333333333)(2048,999.2)(4096,411.1666666666667)(8192,197.66666666666666)
    };
\addlegendentry{Inter-process}
    
\end{semilogxaxis}
\end{tikzpicture}
\end{adjustbox}
\end{subfigure}
}

\subcaptionbox{Sending data with packet loss\label{fig:loss}}
{
\begin{subfigure}[b]{0.3\textwidth}
\begin{adjustbox}{width=\linewidth}
\begin{tikzpicture}[baseline]
\begin{axis}[
    xlabel={Time (s)},
    ylabel={\# of Messages},
    xmin=0, xmax=16,
    ymin=0, ymax=35,
    xtick={2,4,6,8,10,12,14,16},
    ytick={0,5,10,15,20,25,30},
    legend pos=south west,
    grid=both,
    grid style=dashed,
    tick label style={font=\tiny}
]
\node[draw,fill=white] at (4.5,11.0) {0\% packet loss};

\addplot[
    color=blue,
    mark=square,
    ]
    coordinates {
    (1,30.5)(2,29.5)(3,29.0)(4,29.0)(5,29.0)(6,29.0)(7,29.0)(8,29.0)(9,29.0)(10,29.0)(11,29.0)(12,29.0)(13,29.0)(14,29.0)(15,29.0)
    };
\addlegendentry{Messages Sent}

\addplot[
    color=orange,
    mark=square,
    ]
    coordinates {
    (1,29.8)(2,29.6)(3,29.6)(4,29.2)(5,29.6)(6,29.4)(7,29.5)(8,29.9)(9,29.4)(10,29.2)(11,29.7)(12,29.5)(13,29.4)(14,29.6)(15,20.0)
    };

\addlegendentry{Messages Received}

\end{axis}
            \end{tikzpicture}
        \end{adjustbox}
        \end{subfigure}
        \begin{subfigure}[b]{0.3\textwidth}
        \begin{adjustbox}{width=\linewidth}
            \begin{tikzpicture}[baseline]
\begin{axis}[
    xlabel={Time (s)},
    ylabel={\# of Messages},
    xmin=0, xmax=16,
    ymin=0, ymax=35,
    xtick={2,4,6,8,10,12,14,16},
    ytick={0,5,10,15,20,25,30},
    legend pos=south west,
    grid=both,
    grid style=dashed,
    tick label style={font=\tiny}
]
\node[draw,fill=white] at (4.5,11.0) {10\% packet loss};

\addplot[
    color=blue,
    mark=square,
    ]
    coordinates {
    (1,30.6)(2,29.2)(3,29.2)(4,29.0)(5,29.0)(6,29.0)(7,29.0)(8,29.0)(9,29.0)(10,29.0)(11,29.0)(12,29.0)(13,29.0)(14,29.0)(15,29.0)
    };
\addlegendentry{Messages Sent}

\addplot[
    color=orange,
    mark=square,
    ]
    coordinates {
    (1,23.7)(2,25.6)(3,28.3)(4,28.9)(5,29.7)(6,29.2)(7,30.1)(8,27.9)(9,27.5)(10,32.6)(11,29.4)(12,29.4)(13,28.7)(14,30.5)(15,16.7)
    };

\addlegendentry{Messages Received}
    
\end{axis}

\end{tikzpicture}
\end{adjustbox}
\end{subfigure}
\begin{subfigure}[b]{0.3\textwidth}
\begin{adjustbox}{width=\linewidth}
\begin{tikzpicture}[baseline]
\begin{axis}[
    xlabel={Time (s)},
    ylabel={\# of Messages},
    xmin=0, xmax=16,
    ymin=0, ymax=35,
    xtick={2,4,6,8,10,12,14,16},
    ytick={0,5,10,15,20,25,30},
    legend pos=south east,
    grid=both,
    grid style=dashed,
    tick label style={font=\tiny}
]
\node[draw,fill=white] at (9.5,11.0) {20\% packet loss};

\addplot[
    color=blue,
    mark=square,
    ]
    coordinates {
    (1,30.3)(2,29.4)(3,29.3)(4,29.0)(5,29.0)(6,29.0)(7,29.0)(8,29.0)(9,29.0)(10,29.0)(11,29.0)(12,29.0)(13,29.0)(14,29.0)(15,29.0)
    };
\addlegendentry{Messages Sent}

\addplot[
    color=orange,
    mark=square,
    ]
    coordinates {
    (1,7.8)(2,12.5)(3,13.0)(4,13.1)(5,17.8)(6,16.4)(7,22.3)(8,21.4)(9,22.0)(10,26.0)(11,24.6)(12,27.7)(13,25.3)(14,26.8)(15,14.3)
    };

\addlegendentry{Messages Received}
    
\end{axis}
\end{tikzpicture}
\end{adjustbox}
\end{subfigure}
}
\caption{ROS 2 performance results (standard deviation not shown since it is so small)}

\end{figure*}
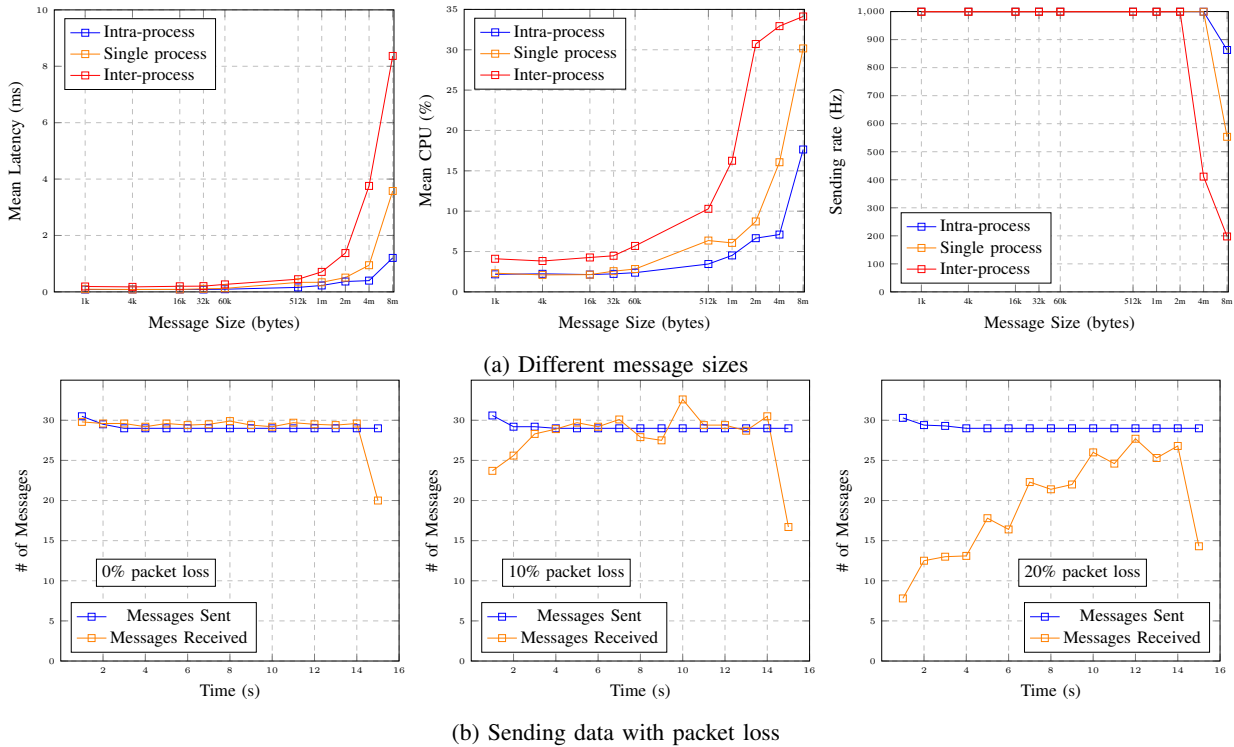

Experiments were conducted to benchmark the networking performance of ROS~2.
The charts in Figure \ref{fig:size} shows the results of sending and receiving different sizes of messages through ROS~2.
This experiment was run on a 6-core Intel i7-6800K CPU running at 3.4GHz, with 32Gb of RAM.
The machine was running Fedora 34, using CycloneDDS, and the latest ROS~2 Rolling distribution packages as of September 23, 2021 \cite{cyclonedds}. The performance tests utilized can be found in \url{https://github.com/ros2/performance_test} and \url{https://github.com/ros2/buildfarm_perf_tests}.
% version 5.13.16

The tests comprise of one publisher and one subscription.
For each message size, 1,000 messages are sent per second and the system records the latency, effective publication rate, and CPU utilization.
The message sizes are selected to test different aspects, ranging from small to larger messages at key intervals.
The test is repeated in different processes, within the same process, and within the same process using intra-process communication. 

The data shows that intra-process communication is the most efficient, with 95th percentile latency below 1 millisecond for all sizes below 8 MB.
Intra-process is the most reliable, meeting the sending rate for all sizes below 8 MB.
This bypasses the middleware stack and delivers data by passing pointers from the publisher to the subscription.
This improvement is particularly magnified when working with large messages, around 1 MB and larger, which are most often associated with images, pointclouds, or other forms of high-resolution data.
When using node composition, the data shows a similar story - the 95th percentile latency is below 1 millisecond with no dropped messages for sizes below  8 MB.

Multi-process communication allows the publisher and subscription to be on separate machines on the network.
Expectedly, it also shows the highest latency, below 1 millisecond until 1 MB, then spiking to 7.85 milliseconds by 8 MB.
The send rate shows a similar trend; at sizes up to 2 MB, the 95th percentile send rate is 1000 Hz, decreasing to 213 Hz for 8 MB.

Using multiple processes and inter-process communication is the most flexible scenario, but it also displays the highest latency and CPU utilization.
Simply using node composition and/or intra-process communication, the latency, CPU utilization, and sending rate are each cut significantly.
However, for small messages, all of the mechanisms were able to publish reliably in excess of 1kHz without loss. 

DDS's default configurations is not particularly effective at communicating information larger than 1 MB, which represents a real challenge to users.
There are a few reasons for this: the small default UDP buffer sizes, UDP fragmentation limits, and DDS reliability guarantees requiring the re-transmit of packets.
Many of these issues can be removed with tuning of the networking parameters at the expense of compute resources.
The performance may also be improved by using the composition and intra-process communication patterns in ROS~2.
Composition is the recommended design pattern in ROS~2 and is made simple to encourage its adoption.

The charts in Figure \ref{fig:loss} shows the resilience of ROS~2 to packet loss in the network.
The tests were run on an Ubuntu 20.04 virtual machine, containing 6 Intel Xeon E5-2666 v3 CPUs at 2.9GHz and 16GB of RAM, using the CycloneDDS {\em rmw}.
For each test the same publisher and subscription node, in separate processes, were run.
The networking was run through {\tt mininet}, a network emulator which lets users specify arbitrary topologies and link characteristics.
In this experiment, the bandwidth was capped at 54Mbps (comparable to a slow wireless network) and the packet loss was varied between 0 and 20 percent.
Each message consisted of an array of 1000 bytes and the number of messages that were received was tallied.
In a network with a moderate amount of loss, ROS~2 can still deliver data over the network effectively.
20\% is a particularly bad networking environment where performance is expected to drop more significantly.

\subsection{Security}

Security is an important element to any modern commercial robotics SDK.
ROS~2 relies on the DDS-Security standard, but also provides an additional suite of tools, SROS2, to make managing security infrastructure easy.
There are 3 main concepts in DDS-security:

\textbf{Authentication} Establishes the identity of a message or participant in the network. ROS~2 uses digital signatures for authentication, known as public key cryptography. SROS2 includes command-line utilities for generating and storing these digital signatures.

\textbf{Access Control} Allows for fine-grained policies to be applied to the authenticated network participants. It allows a participant to only discover approved participants and communicate over pre-approved network interfaces. SROS2 has command-line tools for generating these configurations.

\textbf{Encryption} This ensures that third-parties cannot eavesdrop or replay data into the network. Encryption is performed using AES-GCM symmetric-key cryptography. The key material is derived from the shared secret obtained as part of authentication.

\section{Case Studies}\label{casestudy}
% these should highlight what they are, how they're used, how ROS~2 accelerated these items or general creation of scalable systems
% Summary of each one: ghost on equalizer for a small startup, viper laying the groundwork for future space missions, mission defining a new standard for an industry that's missing it and allowing customers to build custom applications easier, otto multi-robot. Not talk about 4-5 things? Only the main one? Or quick summary on 4-5 points but we have a focus area.
% each case study needs at least 1 quantification / metrics / hard numbers of how ROS2 saves them time / money / resources

Five case studies were conducted that highlight the material acceleration provided by ROS~2.
Each study provides a principally qualitative analysis of ROS~2's influence on each organization based on interviews, customer experiences, and codebases analyzed during the study.
The variety of use cases and scales demonstrates the significance of ROS~2 across the robotics sector.

\subsection{Land: Ghost Robotics}

\begin{figure*}[ht]
     \centering
     \begin{subfigure}[b]{0.49\textwidth}
         \centering
         \includegraphics[trim={0cm 8cm 0cm 6.3cm},clip,height=5.2cm]{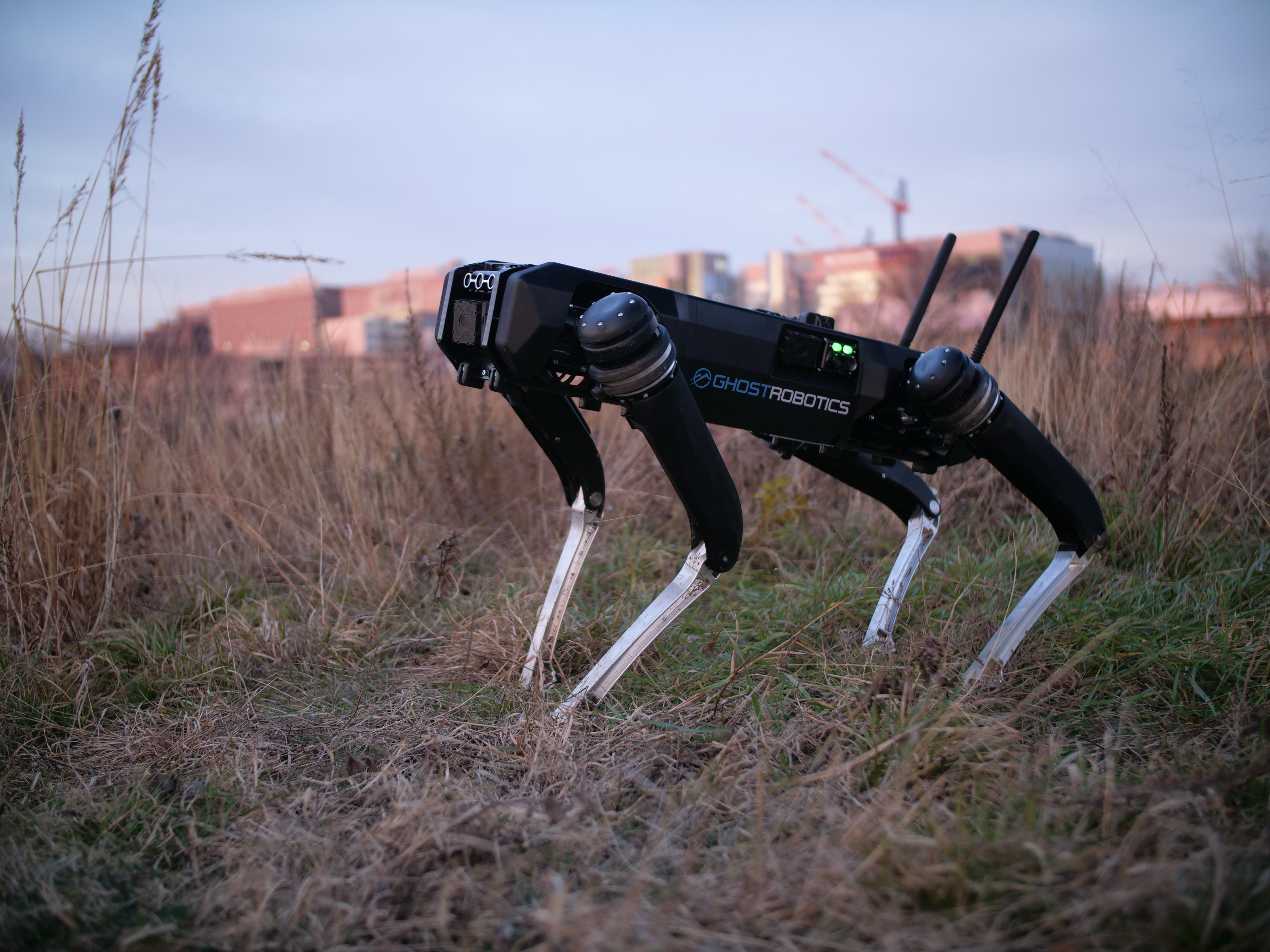}
         \caption{Ghost Robotic's Vision-60 robot.}
         \label{fig:ghost}
     \end{subfigure}
     \hfill
     \begin{subfigure}[b]{0.45\textwidth}
         \centering
         \includegraphics[width=\textwidth]{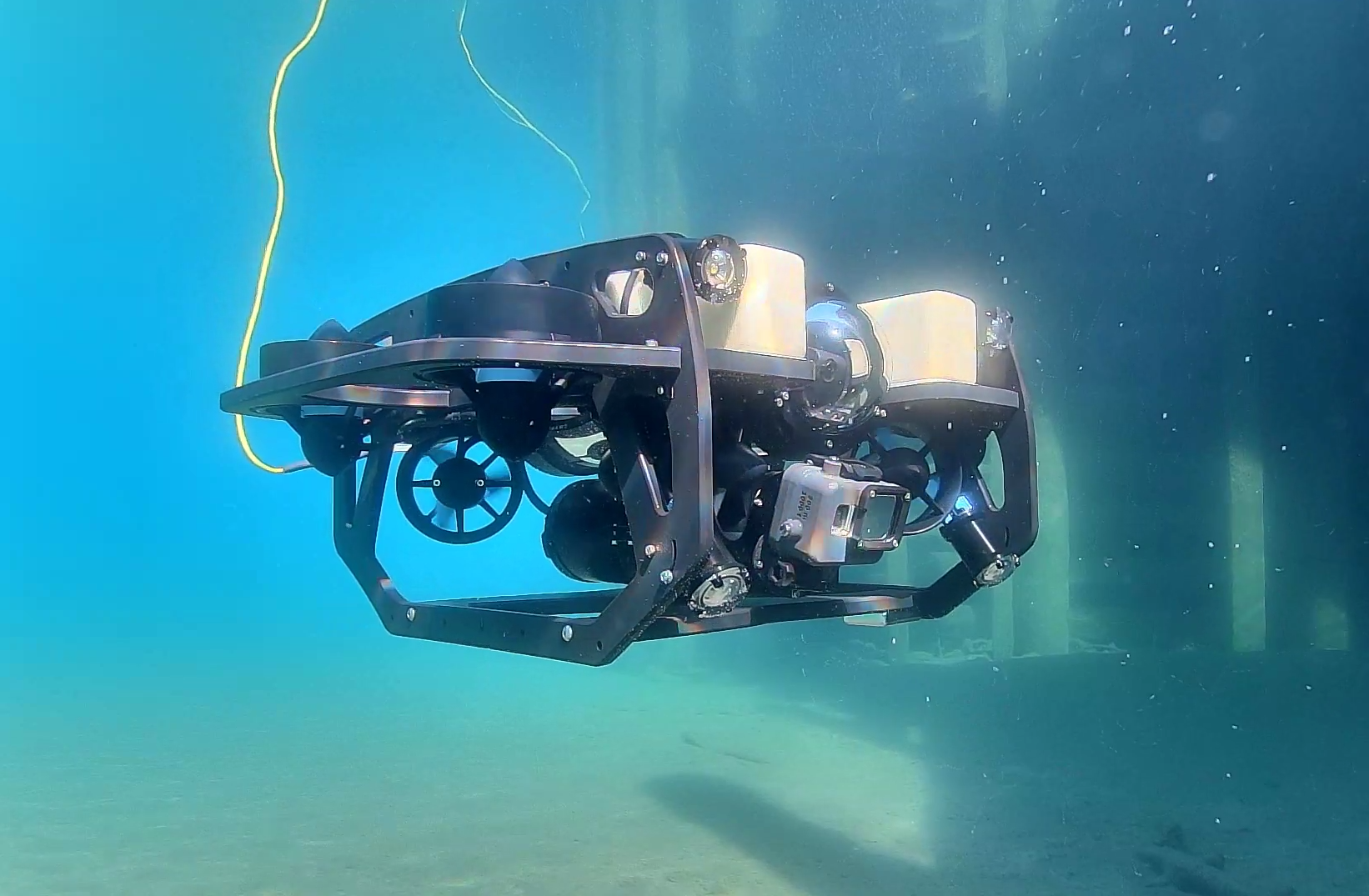}
         \caption{Mission Robotics submersible robot.}
         \label{fig:mission}
     \end{subfigure}
     \begin{subfigure}[b]{0.49\textwidth}
         \centering
         \includegraphics[width=9.42cm]{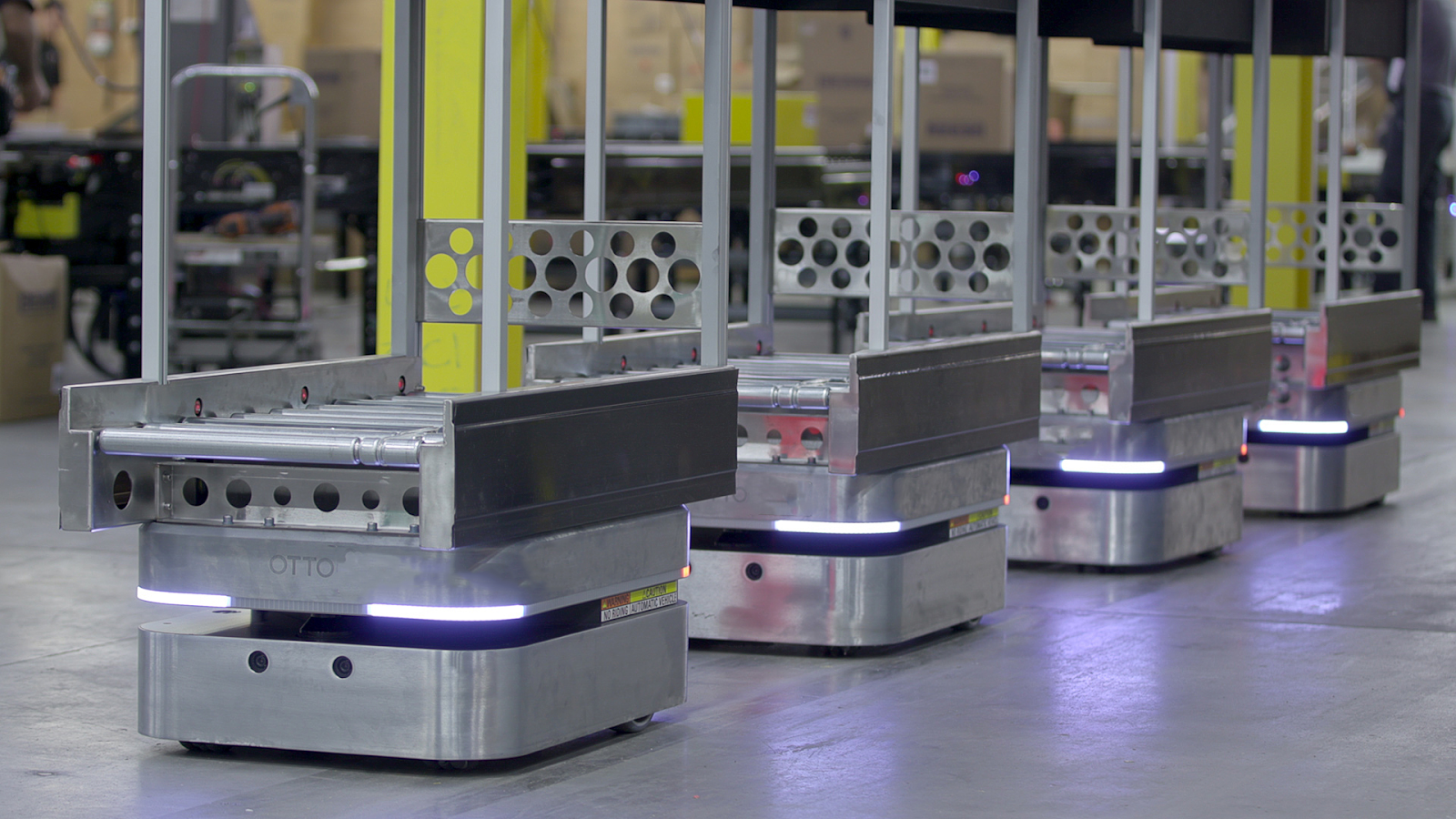}
         \caption{OTTO Motor's `OTTO 100' robots.}
         \label{fig:otto}
     \end{subfigure}
     \hfill
     \begin{subfigure}[b]{0.45\textwidth}
         \centering
         \includegraphics[width=\textwidth]{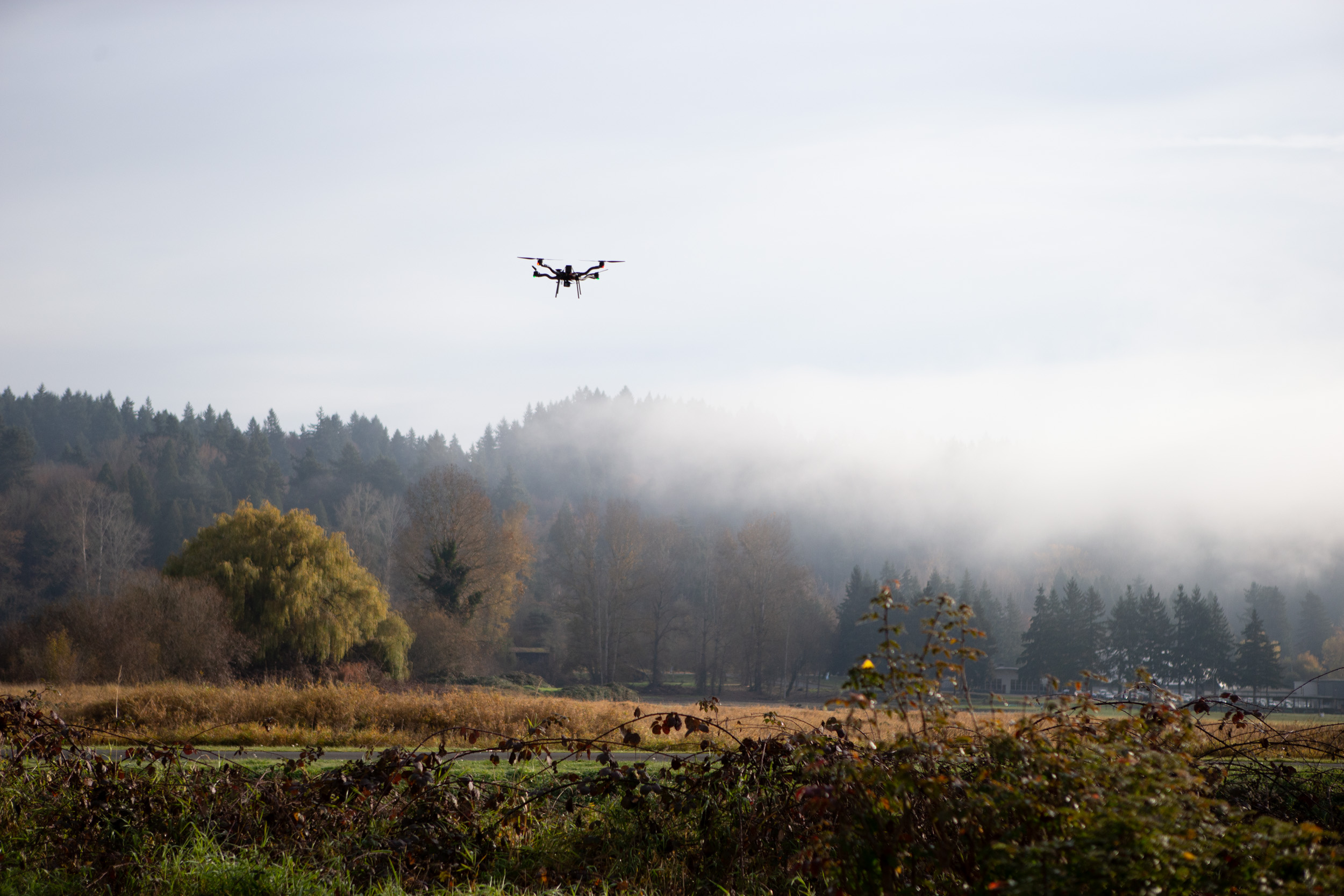}
         \caption{Auterion-powered drone by Freefly.}
         \label{fig:auterion}
     \end{subfigure}
    \caption{Case-study robot systems deployed on land, air, and sea. }
\end{figure*}

Ghost Robotics is a Philadelphia-based company specializing in quadruped robots for defense, enterprise, and research.
Their robots, shown in Figure \ref{fig:ghost}, are made for unstructured natural environments that cannot be traversed by traditional wheeled or tracked robots.
Ghost's Vision-60 robot is being deployed in caves, mines, forests, and deserts, and can easily walk through several inches of water or snow.
Their robots were used by the University of Pennsylvania team in the DARPA Subterranean Challenge and Ghost has active partnerships with the US military for base security and other experimental applications \cite{miller2020tunnel}.

% General overview of ROS use and compute structure
ROS~2 is used on their main compute platform, an Nvidia Jetson Xavier, which handles mission execution, high-level gait planning, terrain mapping, and localization.
Approximately 90\% of Ghost's software utilizes ROS~2 for its communication and architecture, while the remaining is planned to follow suit in the near future.

\subsubsection{Software Architecture}
% Interfaces collaboration
ROS~2 has had a powerful role in structuring their internal collaborations and software design.
Both their high level and mission control software architectures are heavily integrated with ROS~2.
They leverage publish-subscribe interfaces between their main subsystems, allowing them to enjoy a consistent API while the techniques within each are being continuously improved.
This clean separation between projects has allowed them to perform parallel development without disrupting the activities of other teams.

% General Mission control software Architecture description
According to Hunter Allen, Senior Autonomy Engineer, ``It's been great; it is fundamental to our autonomy architecture.''
Their mission control software uses ROS~2 actions to request, cancel, or attain feedback regarding an current mission.
It takes a mission identifier to cross-reference with an internal database of potential missions to execute. %SQLite
Next, it assembles each task in the mission and activates the required capabilities for the particular mission, modeled as lifecycle nodes.
Finally, it executes the mission.

% Explain the component, lifecycle, actions used in the architecture and its significance
Most of Ghost's software is implemented as both lifecycle and component nodes.
The lifecycle node are used to dynamically activate and deactivate features depending on the current mission requirements, such as toggling between GPS-based and VIO-based localization.
They have dozens of unique capabilities readily available for different missions, which take up little background resources when idle.
The component nodes are independent modules developed by multiple teams and combined at run-time.
Ghost found that these strategies are important when collaborating with a large interdisciplinary team on a limited-compute platform.

The provided ROS~2 tools allowed Ghost to create a highly flexible and efficient autonomy system in only a few months.
By contrast, the company estimates it would have taken many years with multiple engineers to create a similar capability if starting from scratch, thereby helping support new custom user applications in the wild.

\subsubsection{The COVID-19 Pandemic}

After the initial COVID-19 lock-downs, the robot software team doubled in effective size while having reduced access to crucial hardware.
At the same time, they were preparing for a demonstration with the US Air Force (USAF) only months away.
Ultimately, the company was successful by pivoting their processes to utilize capabilities made available in ROS~2.

% Gazebo + september 2020 demo + robot access
Prior to the pandemic, the majority of development occurred using robots in their offices.
When access to robots was abruptly stopped, Ghost had to switch development over to the ROS~2 simulator, {\em Gazebo}.
A single engineer was able to create custom Gazebo plugins and simulation files required to represent the quadruped.
This simulation was used to develop the entirety of the USAF demonstration's autonomy system.
This new capability is still utilized long after they were able to return to their offices - it has permitted faster internal development to create custom behaviors and deploy them onto customer's robots.

\subsubsection{ROS~2 as an Equalizer}
% ROS2 as an equalizer against well funded entrenched competitors
ROS~2 is a strong equalizing force for Ghost Robotics.
It has helped them compete effectively with well-funded and entrenched competitors.
Rather than building an end-to-end proprietary portfolio of software, they leverage ROS~2's capabilities where possible.
According to Allen, ``We have a competitive product because we have the tools needed to make a competitive product. We don't have to waste time making what ROS~2 already does.''
With only 23 employees, as of August 2021, compared to competitors an order of magnitude larger, ROS~2 has leveled the playing field.
Ghost was able to release their Vision-60 robot to customers for deployed use after only 30 engineer-years (approx. 7.5 engineers over 4 years).

ROS~2 provides high-quality communications and countless utilities Ghost uses such as {\em TF2, URDF, rosbag, rviz, roscli, Gazebo}, which has accelerated Ghost's robots into the wild.
% The tools provided with ROS~2 have enabled them to build an extensible mission control system on a constrained compute platform and even helped the company weather the COVID-19 pandemic.

\subsection{Sea: Mission Robotics}

Mission Robotics is a San Francisco Bay Area company building marine robots, Figure \ref{fig:mission}.
Their design prioritizes flexibility, supporting a wide range of customers who each customize the platform for their application.
Use cases for Mission's robots include structure inspection, environmental survey, salvage, and security. These tasks are traditionally performed by professional divers, whose time is scarce and valuable. The addition of robots allows important underwater work to be done more often, for longer periods of time, and at far less risk to humans.

Mission's vehicles carry sensors that gather data about the surface and underwater environment.
The robot's sensor suite will vary between users and even between dives performed by a user.
It is important that users be able to add and remove components for a given dive, while being assured of having reliable access to the resulting data.
Mission Robotics uses ROS~2 as the common data bus for these data streams and to enable customers to easily integrate new hardware.

\subsubsection{Customer Architecture}
Mission's core on-robot software does not rely on ROS~2. The engineering team, having experience using DDS, built their internal system on Cyclone and Connext DDS directly \cite{cyclonedds}.
This internal software is maintained exclusively by a small subset of the team at Mission.

The requirements of marine missions are typically specific to customers and not easily generalized, resulting in customization after purchase.
Common practice in the industry is to attach additional sensors or tools as needed, but operate and access each additional peripheral independently, via various device-specific interfaces.

Mission instead uses ROS~2 as the common interface.
When a new sensor, such as a special low-light camera, is to be integrated, a device driver is developed that communicates with the sensor and publishes its data over ROS~2.
The driver is deployed in a Docker container, isolating it from the rest of the vehicle.
Importantly, Mission's customers can create their own extensions, using ROS~2 as the lingua franca, allowing them to modify their robots quickly for custom applications and share common infrastructure.

As an example, Mission worked with Aqualink to add depth sensing to an autonomous surface vessel.
The payload of interest was a Zed stereo camera, which had out-of-the-box ROS~2 drivers, including support for the Jetson Nano single-board computer used.
According to Mission Robotics CTO Charles Cross, ``stereo cameras are growing in popularity in the marine robotics space, especially in clear-water applications like coral reef mapping and species identification.''
By integrating the Zed camera via ROS~2 on the Jetson, Mission and Aqualink were able to create a starting point for anyone wanting to develop new computer vision and autonomy capabilities for marine applications.
This work has attracted the attention of other potential customers, with one of them saying that Mission's approach to payload integration feels ``like it was almost ahead of its time.''

\subsubsection{ROS~2 as an Accelerator}

Support for ROS~2 is a selling point for Mission, providing abstractions for internal systems and offering a familiar developer experience.
Cross reports that for at least three of their customers, ``support for ROS 2 integration has explicitly played a role in their purchasing decision.''

Mission sees ROS~2 as an accelerator for their entire industry. According to Cross, ``in the marine sector, there is little standardization and a lack of building on existing capabilities.'' As a result, ``people keep reinventing the wheel,'' from data logging, to sensor integration, to message formats. This duplicate effort is wasteful and leads to a proliferation of incompatible systems.

Mission Robotics believes that ROS~2 is changing this for marine robotics as it has done for other industries. A common set of messages, APIs, and tools will greatly accelerate the work of Mission and other companies in the sector. In particular, the use of {\em rosbag} for data logging opens the door to collaboration. Such information exchange can benefit robotics engineers, operators, and marine scientists who are the often the end-users of the data.
As Cross says, ``using a consistent communications system is a big win for this industry.''

\subsection{Air: Auterion Systems}

Auterion is an aerial drone startup from Zürich, Switzerland.
It was founded with the goal of nurturing the open-source PX4 Autopilot developer community \cite{px4}.
Building on their PX4 flight experience, Auterion produces a commercial autopilot based on the project and offers commercial support for customer integrations.
Auterion's products are used widely throughout the industry and support many types of airframes, including Freefly shown in Fig. \ref{fig:auterion}.

Historically, drones could only be operated safely by skilled pilots or in open spaces.
Auterion aims to bring drones into unstructured spaces with hazards while operating under more autonomy.
With their emphasis on open standards, Auterion selected ROS~2 to integrate higher level functionality into their drone systems alongside the PX4 Autopilot.

\subsubsection{Logging and Introspection}

ROS~2's logging, introspection, and debugging improved the efficiency of their development process on {\em Skynode}, their fully-integrated autopilot solution.
The logging capabilities from ROS~2 are used to collect run-time events such as errors, debug outputs, and other metadata about the system.
These are stored for later analysis and debugging.
Auterion also relies on {\em rosbag2} to collect the raw data stream at run-time from all layers of the system, from sensor streams to vehicle behaviors.
This comprehensive logging is especially valuable for drones because environmental aspects such as wind have important effects on flight conditions, which are difficult to reproduce.
As a consequence, ROS~2's dataset and logging capabilities are central to effective development, debugging, and validation processes.

Auterion also takes advantage of robust introspection capabilities.
Auterion uses {\em rviz2}, a 3-dimensional renderer which can visualize drones and all of their sensor data in an interactive environment.

The 3-dimensional visualisation, data recording, and logging capabilities in ROS~2 were one of the driving reasons Auterion utilizes ROS~2.
The value of these tools was captured most succinctly by Nuno Marques, a Software Engineer at Auterion, ``The fact that we have introspection and visualization tools make all the difference.''
Leveraging these capabilities has allowed the company to focus their development efforts on core flight control capabilities and customer requirements rather than building foundational tooling.

\subsubsection{Safe, Automated Testing}

Flying drones has inherent risks to people and things on the ground, as well as to the airframe itself.
A great deal of labor and time is required to conduct safe flight testing since every physical flight has a risk of crashing.
In simulation, however, the cost and risks associated with test flights is near zero.
A failure in simulation can be fixed and iterated upon quickly, then rerun.
Auterion uses ROS~2's simulation, {\em Gazebo} to be able to conduct end-to-end tests of the software prior to hardware testing to validate safe functionality.

Gazebo is used in their continuous integration pipeline to prevent regressions on an array of vehicle types and scenarios.
Tests are run in parallel for fast results, which allows developers to focus on a specific problem while remaining confident the software is safe.

Auterion also leverages simulation testing to validate features in challenging scenarios during development.
For example, they can setup flight regimes or specific situations which are important to validate their work.
In 2021, Auterion flew approximately 22,000 hours within Gazebo, including high-risk scenarios impractical to test with hardware.
Auterion estimates that these simulations replaced 12 full-time engineers to provide the same value in live tests.
Since the cost of their airframes range from \$1,000 to \$100,000, there is considerable risk in any testing - especially in dangerous flight conditions which need testing.
ROS~2 simulations in development and validation combine to enable lower costs and faster development.

\subsection{Space: NASA VIPER}
\begin{figure*}[ht]
\centering
\begin{subfigure}{.5\textwidth}
    \centering
    \includegraphics[width=\textwidth]{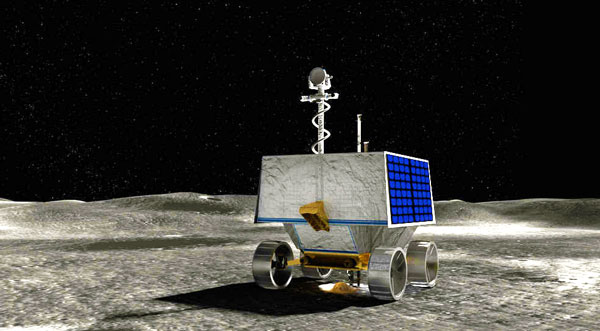}
    \caption{}
    \label{fig:viper_render}
\end{subfigure}%
\begin{subfigure}{.5\textwidth}
    \centering
    \includegraphics[width=\textwidth]{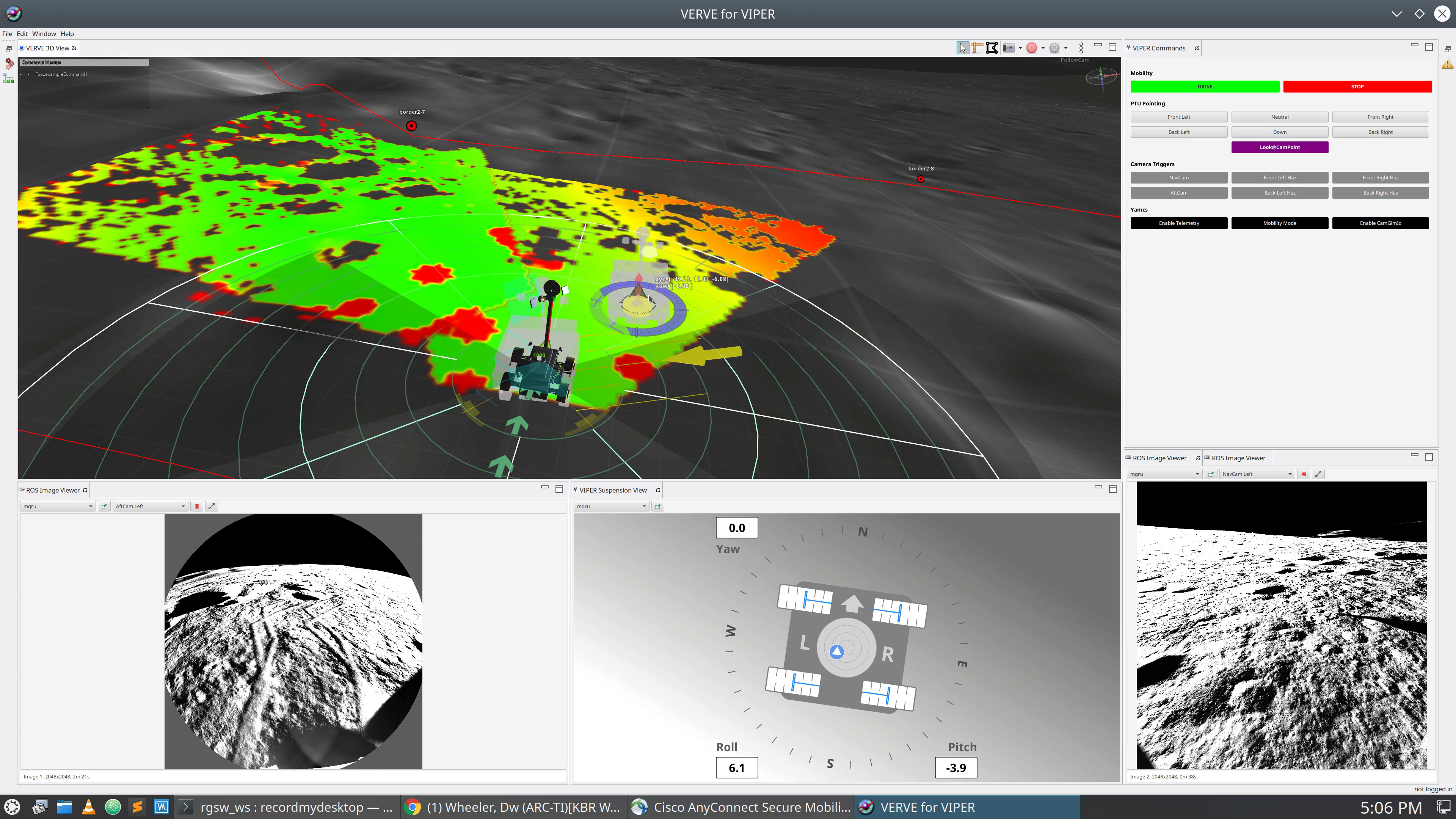}
    \caption{}
    \label{fig:viper_command}
\end{subfigure}
\caption{(a) VIPER on Lunar Surface (rendering), (b) Command and Operations Software. }
\label{fig:viper}
\end{figure*}

NASA's Volatiles Investigating Polar Exploration Rover (VIPER) mission is scheduled to be launched to the southern polar region of the Moon in November 2023.
The VIPER rover will use a variety of instruments to search for water ice and other resources during a 100-day mission.
Earth compute resources will be used to map, register terrain, and compute stereo solutions to aid in operations through its X-band link to the Deep Space Network.
Many of the Earth-based operation tools, compute modules, and high-fidelity simulations are based on ROS~2 and Gazebo, as shown in Figure \ref{fig:viper}.

NASA's Core Flight system provides hardware interfaces, basic error checking, and payload services \cite{nasacfs}.
A satellite link carries commands to and telemetry from the rover.
Earth-bound telemetry is received and sent to a ROS~2 network and processed by an ensemble of nodes.
The nodes transform the image data into pointclouds, compute visual odometry and terrain registration, and fuses the data to provide pose corrections.
That data is fed into NASA's Visual Environment for Remote Virtual Exploration (VERVE), which allows operators to visualize the rover's environment \cite{nasaverve}.
The operators use the result to simulate a move, and then finally execute the move on the rover.

\subsubsection{Mission Testing in Simulation}

Since VIPER is a spaceflight mission, the team is focused on producing highly reliably software.
To achieve this, they are extensively utilizing Gazebo to provide high-fidelity testing of all their components and systems.
Mark Allen said ``having a simulator [Gazebo] is essential for the development of all the VIPER software in some capacity.''

The VIPER team turned to Gazebo to aid in development since it was infeasible to model an accurately functioning lunar rover on Earth.
They emphasized ``the Lunar environment is so unique, with lighting and gravity, testing in simulation [is] incredibly important since its impossible to test on the ground on Earth effectively.''
The project was able to create a simulation utilizing custom plugins to Gazebo's user interfaces.
It is designed for a high degree of customization to support a broad range of robotics needs - even space.

NASA developed new plugins to model mission-specifics, such as camera lens flare, lunar lighting conditions, gravity, and terrain on the lunar surface.
NASA was able to simulate the vehicle interfaces down to low-level serial links.
The simulation was valuable to help iterate and improve upon system design choices for VIPER.
With the rover simulated down to the hardware level, the VIPER team used Gazebo to test and validate almost all of their rover's software prior to launch.

VIPER reused 284,500 significant lines of code (SLOC) without modification from Gazebo, modifying $<1\%$ to pass validation.
NASA's estimated development rate for the simulator was 116 SLOC per work-month (2456 work-months to fully implement).
This code reuse accelerated development allowing them to produce a simulation in merely 266 work-months focused VIPER specific elements \cite{nasavipercost}.

A combination of Gazebo and ROS~2 is used to train the rover's operators.
ROS~2 is used to inject faults into the rover; using VERVE, the operators need to determine how to clear the faults to get the rover moving.

\subsubsection{Creating a Legacy}

NASA has utilized many different communication mechanisms but in recent years, many projects have chosen DDS because of its ability to traverse satellite links that may have high latency, low bandwidth, and low reliability.
The VIPER team evaluated the options and selected DDS as well for the Earth-based operations.

Besides a communications mechanism, the VIPER team was eager to use ROS~2 for its rapid development capabilities, introspection and visualization tools, and openly available source.
These characteristics shorten the learning curve for new engineers to apply what they know onto flight missions.

However, using new software in a flight mission requires a rigorous Verification and Validation (V\&V) process.
NASA prefers to use components that has been vetted in previous missions; leveraging heritage software leads to reduced development times and costs \cite{nasaheritage}.
VIPER is reusing 84\% of the 588,000 lines of code from the Resource Prospector along with Gazebo and approximately 312 open-source ROS~2 packages \cite{nasavipercost}.
ROS~2 has not been used in prior missions, but the VIPER team decided that the features that it provides was worth the extra administrative overhead of going through the process.

After ROS~2 has been validated and used in ground operations for the VIPER mission, it becomes much easier for ROS~2 to be used in future missions in multiple roles and allow for more reuse of robotic software between mission programs.

\subsection{Large Scale: OTTO Motors}
OTTO Motors is an Ontario-based Clearpath Robotics spinoff company selling land and sea research platforms.
OTTO produces warehouse and factory material handling services using autonomous robots to replace manually controlled equipment at scale - Figure \ref{fig:otto}.
They have deployed thousands of robots worldwide and operate fleets of over 100 in a single facility.
Customers such as Toyota and General Electric have adopted OTTO.
This case study provides unique insight into large scale robot applications.
ROS~2 has coordinated more than 2 million hours of operation and 1.5 million kilometers traveled since it was deployed on OTTO's robots.

\subsubsection{Scaling Multi-Robot Technology}
OTTO Motors originally developed their technology on ROS~1.
Using it, they were unable to test more than 25 robots on the same shared ROS~1 network using a custom multi-master system with their fleet management software.
This was sufficient for small fleets, but as OTTO grew into larger facilities, this became a bottleneck.

OTTO conducted a survey of available technologies and independently came to the same conclusion that the best technology for the multi-robot fleet communications was DDS.
The greater network effect worked in their favor to continue in the ROS ecosystem, thus they were one of the early adopters of ROS~2.
This allowed them to take advantage of the capabilities enabled in ROS, while not independently maintaining a proprietary DDS framework.

After migrating to ROS~2, OTTO was able to scale up to 100+ robots in customer facilities.
Larger multi-robot scale was enabled because ROS~2 has fine-grained and scalable network topology management as well as better support for managing bandwidth through QoS on shared network links.
These deployments in 2017 represented some of the first commercial deployments of ROS~2 anywhere in the world.
ROS~2 demonstrably accelerated their time to market by quickly enabling them to scale to unprecedented numbers of robots.
OTTO Motors estimates they have saved between \$1M to \$5M over 5 years by using ROS~2.
In addition they saved hundreds of engineering hours by not rewriting these tools into a proprietary framework.

OTTO Motor's CTO Ryan Gariepy considers the ROS ecosystem to be necessary to the business, ``Had ROS writ large not existed, the whole business might not have been feasible. It would have been too expensive.''
He estimates that their continuing engineering costs would be 5--10\% higher annually without it.

\subsubsection{Acceleration of Development}

OTTO Motors' development and deployment have been sped up in two additional areas.
First, it has accelerated their internal feature development process.
The distributed architecture and isolation of processes have allowed a large, physically distributed team to collaborate.
Using clearly defined ROS~2 interfaces allowed OTTO to separate major classes of tasks.
Ryan Gariepy stated in an interview, ``at the scale of robots we're building and the complexity that is modern manufacturing, you really need the flexibility to patch in and out capabilities and share across a large team.''
Their product software is spread across many repositories owned by different teams in a diverse set of languages, combined at run-time via ROS~2.

Next, providing ROS~2 support has proven valuable to their customers and clients.
OTTO and Clearpath sell their platforms to other businesses to build on top of for custom products.
A company recently bought platforms from OTTO to create UV sanitizing robots in response to the COVID-19 pandemic.
Since they have both clearly defined and standard APIs, these external collaborators easily leveraged the robot platform and tied it into their autonomy systems.
Ryan Gariepy summarized it as follows, ``With the ROS APIs we provide, our external partners are now able to build apps on top of our autonomy capabilities without requiring us to train them in robotics concepts or proprietary libraries.''
This ability to separate concerns and abstract vendor specific hardware, even an entire robot platform, allows companies to rapidly build a new product to ensure public safety.

\subsection{Discussion}

These five case studies illustrate an extensive range of applications, environments, and rationales for the use of ROS~2.
These were selected to provide a unique cross-section of modern, applied robotics systems deployed in every domain.
However different they are in their applications, there exists several common threads they share.

ROS~2 enables many to better {\em reuse} software components in their systems.
Mission Robotics leverages the ROS community's device drivers and integrations such that their customers can quickly adapt to a particular use-case in marine robotics.
Likewise, Auterion uses not only lower level drivers but also the higher level algorithms from the community.
The VIPER team uses ROS~2 to facilitate software reuse within the agency.
During our interviews, they expressed that it was challenging to get other NASA groups to reuse code and that the ROS ecosystem has internal name recognition making it easier to encourage such collaborations.

Another common thread was enabling {\em collaborations}, both internally and externally.
Ghost Robotics and OTTO Motors use interfaces and composition nodes to separate parts of a complex system so teams can collaborate without needing to concern themselves with the details of other parts of the system.
Both Mission and Auterion are able to build custom solutions collaboratively with their customers by utilizing ROS~2. 

Finally, ROS~2 has allowed businesses to accelerate others via the sale of {\em trusted platforms}.
All of the companies surveyed sell their platforms to other businesses to build products on top of.
The proliferation of ROS expertise in the industry, matched with its freely available licensing, has made it the major robotics SDK.
By employing ROS~2 and its conventions, they are able to sell platforms that can be put to work in bespoke applications quickly.

% some closing statement in the section
It should be noted that these themes, software reuse; collaborations; trusted platforms, are highly correlated with the design principles laid out in Section \ref{sec:principles}.
In particular, they are in line with the design principles of Distribution, Abstraction, and Modularity.
The adherence to those design principles have directly resulted in the emergent themes in our studies, which represent some of the largest acceleration factors for the robotics industry today.

\section{Conclusion}

ROS~2 has been redesigned from the ground up to meet the challenges of modern robotics.
It was designed based off of a thoughtful set of principles, modern robotics requirements, and support for extensive customization.
Largely based on DDS, ROS~2 is a reliable and high quality robotics framework that can support a broad range of applications.
This framework continues to help accelerate the deployment of robots out of the lab, into the wild, and is driving the next wave of the robotics revolution.

We have shown through a series of case studies how it is demonstrably accelerating companies and institutions into useful deployment in many types of environments at a wide variety of scales.
They display that ROS~2 is an enabler, an equalizer, and an accelerator.
The standardization around ROS~2 in a variety of industries is creating opportunities for new collaborations, faster development, and propelling newly developed technologies forward.
This trend will likely continue to manifest in the coming years as ROS~2 continues to reach its peak maturity.

\section{Acknowledgments}

We would like to thank the companies representatives interviewed in the case studies.
This includes: Hunter Allen and James Laney from Ghost Robotics, Charles Cross from Mission Robotics, Nuno Marques and Markus Achtelik from Auterion, Mark Allen and Terry Fong from NASA Ames, and Ryan Gariepy from OTTO Motors.
We would also like to thank the team at Open Robotics, members of the ROS~2 Technical Steering Committee, and the community for their  passionate support.

\bibliography{refs}

\begin{thebibliography}{10}

\bibitem{ros2009icra}
Morgan Quigley, Brian Gerkey, Ken Conley, Josh Faust, Tully Foote, Jeremy
  Leibs, Eric Berger, Rob Wheeler, and Andrew Ng.
\newblock {ROS: an open-source Robot Operating System}.
\newblock In {\em IEEE International Conference on Robotics and Automation
  Workshop on Open Source Software}, 2009.

\bibitem{Chitta2017}
Sachin Chitta, Eitan Marder-Eppstein, Wim Meeussen, Vijay Pradeep,
  Adolfo~Rodríguez Tsouroukdissian, Jonathan Bohren, David Coleman, Bence
  Magyar, Gennaro Raiola, Mathias Lüdtke, and Enrique~Fernandez Perdomo.
\newblock {ros\_control: A generic and simple control framework for ROS}.
\newblock {\em Journal of Open Source Software}, 2(20):456, 2017.

\bibitem{5509725}
Eitan Marder-Eppstein, Eric Berger, Tully Foote, Brian Gerkey, and Kurt
  Konolige.
\newblock {The Office Marathon: Robust navigation in an indoor office
  environment}.
\newblock In {\em IEEE International Conference on Robotics and Automation},
  pages 300--307, 2010.

\bibitem{moveit}
David Coleman, Ioan Sucan, Sachin Chitta, and Nikolaus Correll.
\newblock {Reducing the Barrier to Entry of Complex Robotic Software: a MoveIt!
  Case Study}.
\newblock {\em Journal of Software Engineering for Robotics}, 5(1):3–16,
  2014.

\bibitem{Cairl2018}
Brian Cairl (Fetch~Robotics Inc.).
\newblock Deterministic, asynchronous message driven task execution with ros.
\newblock In {\em ROSCon Madrid 2018}. Open Robotics, September 2018.

\bibitem{macenski2020marathon2}
Steve Macenski, Francisco Martín, Ruffin White, and Jonatan Ginés~Clavero.
\newblock {The Marathon 2: A Navigation System}.
\newblock In {\em IEEE/RSJ International Conference on Intelligent Robots and
  Systems}, 2020.

\bibitem{1203555}
G.~Pardo-Castellote.
\newblock {OMG Data-Distribution Service: architectural overview}.
\newblock In {\em International Conference on Distributed Computing Systems
  Workshops}, pages 200--206, 2003.

\bibitem{ros2-on-dds}
{William Woodall}.
\newblock {ROS on DDS}.
\newblock \url{https://design.ros2.org/articles/ros_on_dds.html}, accessed
  February 11, 2022.

\bibitem{kuipers2017}
Benjamin Kuipers, Edward~A. Feigenbaum, Peter~E. Hart, and Nils~J. Nilsson.
\newblock {Shakey: From Conception to History}.
\newblock {\em AI Magazine}, pages 88--103, 2017.

\bibitem{fikes1971strips}
Richard~E Fikes and Nils~J Nilsson.
\newblock Strips: A new approach to the application of theorem proving to
  problem solving.
\newblock {\em Artificial intelligence}, 2(3-4):189--208, 1971.

\bibitem{brooks1986robust}
Rodney Brooks.
\newblock A robust layered control system for a mobile robot.
\newblock {\em IEEE journal on robotics and automation}, 2(1):14--23, 1986.

\bibitem{gat1998three}
Erann Gat, R~Peter Bonnasso, Robin Murphy, et~al.
\newblock On three-layer architectures.
\newblock {\em Artificial intelligence and mobile robots}, 195:210, 1998.

\bibitem{simmons1994structured}
Reid~G Simmons.
\newblock Structured control for autonomous robots.
\newblock {\em IEEE transactions on robotics and automation}, 10(1):34--43,
  1994.

\bibitem{montemerlo2003perspectives}
Michael Montemerlo, Nicholas Roy, and Sebastian Thrun.
\newblock Perspectives on standardization in mobile robot programming: The
  carnegie mellon navigation (carmen) toolkit.
\newblock In {\em Proceedings 2003 IEEE/RSJ International Conference on
  Intelligent Robots and Systems (IROS 2003)(Cat. No. 03CH37453)}, volume~3,
  pages 2436--2441. IEEE, 2003.

\bibitem{mohan1994recent}
C~Mohan and R~Dievendorff.
\newblock Recent work on distributed commit protocols, and recoverable
  messaging and queuing.
\newblock {\em Data Engineering}, 17(1):1, 1994.

\bibitem{waldo1999jini}
Jim Waldo.
\newblock The jini architecture for network-centric computing.
\newblock {\em Communications of the ACM}, 42(7):76--82, 1999.

\bibitem{mqtt2019}
OASIS.
\newblock {\em {MQTT Version 5.0: OASIS Standard}}, 2019.

\bibitem{player2001iros}
Brian~P. Gerkey, Richard~T. Vaughan, Kasper Støy, Andrew Howard, Gaurav~S.
  Sukhatme, and Maja~J Matarić.
\newblock {Most Valuable Player: A Robot Device Server for Distributed
  Control}.
\newblock In {\em IEEE/RSJ International Conference on Intelligent Robots and
  Systems}, 2001.

\bibitem{yarp2006ars}
Giorgio Metta, Paul Fitzpatrick, and Lorenzo Natale.
\newblock {YARP: Yet Another Robot Platform}.
\newblock {\em International Journal of Advanced Robotic Systems}, 3(1):43--48,
  2006.

\bibitem{lcm2010iros}
Albert~S. Huang, Edwin Olson, and David~C. Moore.
\newblock {LCM: Lightweight Communications and Marshalling}.
\newblock In {\em IEEE/RSJ International Conference on Intelligent Robots and
  Systems}, pages 4057--4062, 2010.

\bibitem{orocos2003icra}
H.~Bruyninckx, P.~Soetens, and B.~Koninckx.
\newblock {The real-time motion control core of the Orocos project}.
\newblock In {\em IEEE International Conference on Robotics and Automation},
  2003.

\bibitem{apache2license}
Apache~Software Foundation.
\newblock {Apache License, Version 2.0}.
\newblock \url{https://www.apache.org/licenses/LICENSE-2.0.html}, accessed
  September 3, 2021.

\bibitem{Birman1987ExploitingVS}
K.~Birman and T.~A. Joseph.
\newblock {Exploiting virtual synchrony in distributed systems}.
\newblock In {\em {ACM Symposium on Operating Systems Principles}}, pages
  123--138, 1987.

\bibitem{corbet2005linux}
Jonathan Corbet, Alessandro Rubini, and Greg Kroah-Hartman.
\newblock {\em Linux device drivers}.
\newblock "O'Reilly Media, Inc.", 2005.

\bibitem{muhl2006distributed}
Gero M{\"u}hl, Ludger Fiege, and Peter Pietzuch.
\newblock {\em Distributed event-based systems}.
\newblock Springer Science \& Business Media, 2006.

\bibitem{McIlroy78}
M.~D. McIlroy, E.~N. Pinson, and B.~A. Tague.
\newblock {Unix Time-Sharing System: Foreword}.
\newblock {\em {The Bell System Technical Journal}}, 57(6):1899--1904, 1978.

\bibitem{Gerard2017}
Aravind Sundaresan and Leonard Gerard.
\newblock Secure ros: Imposing secure communication in a ros system.
\newblock In {\em ROSCon Vancouver 2017}. Open Robotics, September 2017.

\bibitem{ros2-dds-security}
{Kyle Fazzari}.
\newblock {ROS 2 DDS-Security integration}.
\newblock \url{https://design.ros2.org/articles/ros2_dds_security.html},
  accessed September 6, 2021.

\bibitem{ddssecurity}
{OMG}.
\newblock {DDS Security}.
\newblock \url{https://www.omg.org/spec/DDS-SECURITY/1.0/PDF}, accessed
  February 9, 2022.

\bibitem{White2018}
Ruffin White, Gianluca Caiazza, Henrik Christensen, and Agostino Cortesi.
\newblock Procedurally provisioned access control for robotic systems.
\newblock In {\em IEEE/RSJ International Conference on Intelligent Robots and
  Systems}, 2018.

\bibitem{Lutkebohle2019}
Ingo Lütkebohle, Borja~Outerelo Gamarra, Iñigo~Muguruza Goenaga, Jaime~Martin
  Losa, and Víctor~Mayoral Vilches.
\newblock {micro-ROS: ROS 2 on microcontrollers}.
\newblock In {\em ROSCon}. Open Robotics, October 2019.

\bibitem{dds-spec}
{Object Management Group}.
\newblock {Data Distribution Service for Real-time Systems Specification},
  December, 2004.

\bibitem{puck2020distributed}
Lennart Puck, P~Keller, Tristan Schnell, Carsten Plasberg, Atanas Tanev, Georg
  Heppner, Arne R{\"o}nnau, and R{\"u}diger Dillmann.
\newblock {Distributed and Synchronized Setup towards Real-Time Robotic Control
  using ROS2 on Linux}.
\newblock In {\em 2020 IEEE 16th International Conference on Automation Science
  and Engineering (CASE)}, pages 1287--1293. IEEE, 2020.

\bibitem{staschulat2020rclc}
Jan Staschulat, Ingo L{\"u}tkebohle, and Ralph Lange.
\newblock {The rclc Executor: Domain-specific deterministic scheduling
  mechanisms for ROS applications on microcontrollers: work-in-progress}.
\newblock In {\em International Conference on Embedded Software}, pages 18--19.
  IEEE, 2020.

\bibitem{apex-cert}
Mehul Sagar.
\newblock {ISO Certification of ROS 2}.
\newblock In {\em Embedded World Conference}, March 2021.

\bibitem{asio}
{John Torjo}.
\newblock {Asio C++ Network Programming: Enhance Your Skills with Practical
  Examples for C++ Network Programming}, 2013.

\bibitem{aws-iot-roborunner}
{Channy Yun}.
\newblock {AWS IoT RoboRunner for Building Robot Fleet Management
  Applications}, accessed February 11, 2022.

\bibitem{ros2design-site}
{ROS 2 Design}.
\newblock \url{http://design.ros2.org/}, accessed August 5, 2021.

\bibitem{reps-site}
{ROS Enhancement Proposals}.
\newblock \url{https://ros.org/reps/rep-0000.html}, accessed August 5, 2021.

\bibitem{rep-2004}
{William Woodall}.
\newblock {REP 2004: Package Quality Categories}.
\newblock \url{https://ros.org/reps/rep-2004.html}, accessed August 5, 2021.

\bibitem{cyclonedds}
Eclipse Foundation.
\newblock Cyclone \uppercase{DDS}.
\newblock \url{https://cyclonedds.io/}, accessed September 3, 2021.

\bibitem{miller2020tunnel}
Ian~D. Miller, Fernando Cladera, Anthony Cowley, Shreyas~S. Shivakumar,
  Elijah~S. Lee, Laura Jarin-Lipschitz, Akhilesh Bhat, Neil Rodrigues, Alex
  Zhou, Avraham Cohen, Adarsh Kulkarni, James Laney, Camillo~Jose Taylor, and
  Vijay Kumar.
\newblock Mine tunnel exploration using multiple quadrupedal robots, 2020.

\bibitem{px4}
Lorenz Meier, Dominik Honegger, and Marc Pollefeys.
\newblock Px4: A node-based multithreaded open source robotics framework for
  deeply embedded platforms.
\newblock In {\em 2015 IEEE International Conference on Robotics and Automation
  (ICRA)}, pages 6235--6240, 2015.

\bibitem{nasacfs}
{David McComas}.
\newblock {NASA/GSFC’s Flight Software Core Flight System}.
\newblock In {\em {Flight Software Workshop}}, {2012}.

\bibitem{nasaverve}
Susan~Y Lee, David Lees, Tamar Cohen, Mark Allan, Matthew Deans, Theodore
  Morse, Eric Park, and Trey Smith.
\newblock Reusable science tools for analog exploration missions: xgds web
  tools, verve, and gigapan voyage.
\newblock {\em Acta Astronautica}, 90(2):268--288, 2013.

\bibitem{nasavipercost}
Sherry Stukes, Mark Allan, Matthew~Deans Georgia~Bajjalieh, Terrence Fong,
  Jairus Hihn, and Hans Utz.
\newblock An innovative approach to modeling viper rover software life cycle
  cost.
\newblock In {\em 2021 IEEE Aerospace Conference (50100)}. IEEE, 2021.

\bibitem{nasaheritage}
{Charley Price}.
\newblock {Heritage Software Save up to 97\% on future V\&V for real projects}.
\newblock
  \url{https://www.nasa.gov/sites/default/files/03-09_ivv_guidance_for_ivv_for_product_line_software.pdf},
  accessed September 7, 2021.

\end{thebibliography}
\bibliographystyle{unsrt}

% \begin{thebibliography}{00}
% \input{main}
% \end{thebibliography}

\end{document}